\documentclass[runningheads,table]{llncs}
\usepackage{tabularx}
\usepackage{booktabs}
\usepackage{cite}
\usepackage{amsmath,amssymb,amsfonts}
\usepackage{algorithmic}
\usepackage{textcomp}
\usepackage{graphicx}
\usepackage{subcaption}
\usepackage{caption}
\usepackage{hyperref}
\usepackage{array}
\usepackage{colortbl}
\usepackage[table]{xcolor}

\definecolor{mygray}{gray}{0.6}
\definecolor{mylightgray}{gray}{0.9}

\definecolor{headercolor}{gray}{0.85}

\begin{document}
\title{Enhancing the Accuracy of Predictors of Activity Sequences of Business Processes\thanks{Work funded by the European Research Council (PIX Project).}}
\titlerunning{Daemon Action Approach for Activity Sequence Prediction}
%
\author{Muhammad Awais Ali \and
Marlon Dumas \and
Fredrik Milani}
\authorrunning{Muhammad Awais Ali et al.}
%
\institute{University of Tartu, Tartu, Estonia \\
\email{\{firstname.lastname\}@ut.ee}
}

\maketitle              
\begin{abstract}
Predictive process monitoring is an evolving research field that studies how to train and use predictive models for operational decision-making. One of the problems studied in this field is that of predicting the sequence of upcoming activities in a case up to its completion, a.k.a. the case suffix. The prediction of case suffixes provides input to estimate short-term workloads and execution times under different resource schedules. Existing methods to address this problem often generate suffixes wherein some activities are repeated many times, whereas this pattern is not observed in the data. Closer examination shows that this shortcoming stems from the approach used to sample the successive activity instances to generate a case suffix. Accordingly, the paper introduces a sampling approach aimed at reducing repetitions of activities in the predicted case suffixes. The approach, namely Deamon action, strikes a balance between exploration and exploitation when generating the successive activity instances. We enhance a deep learning approach for case suffix predictions using this sampling approach, and experimentally show that the enhanced approach outperforms the unenhanced ones with respect to both control-flow and temporal accuracy measures.
\end{abstract}

\section{Introduction}

Predictive process monitoring\cite{TaxVRD17} is a set of techniques to predict future states or properties of ongoing cases of a business process. These techniques are based on event logs extracted from information systems\cite{DumasRMR18}. These predictions\cite{KubrakMND23} help operational managers to make decisions to improve the performance of the process, such as reallocating resources from one case to another to increase the probability of achieving positive case outcomes or to reduce delays~\cite{ToosinezhadFKA20}. 

Existing predictive process monitoring techniques address a range of prediction targets\cite{RamaManeiroVL23}, such as predicting the outcome of a case\cite{TeinemaaDRM19,KratschMRS21}, its remaining cycle time\cite{VerenichDRMT19}, the next activity instance\cite{TaxVRD17}, and the sequence of activity instances until completion of the case\cite{TaxVRD17,0001DR19,TaymouriREBV20}. In this paper, we are concerned with the latter prediction target. Given an ongoing case of a process, and given the sequence of activity instances that have occurred up to that point in time (a \emph{case prefix}), the goal is to predictive the sequence of activity instances up to completion of the case (herein called the \emph{case suffix} for short). For each activity instance in this sequence, we specifically aim to predict its activity type and its completion time. Thus, incidentally, the paper also tackles the problem of predicting the remaining time until completion of the case, since this remaining time is equal to the difference between the completion time of the last activity instance in the case suffix and the completion time of the last activity instance in the given case prefix.

A number of methods for case suffix prediction have been proposed in the literature~\cite{0001DR19,TaxVRD17,Francescomarino17,KendallGC18,LinWW19}. A common idea behind these approaches is to train a machine learning model (usually with a deep learning architecture) on case prefixes extracted from historical process execution data (\emph{event logs}). Given a case prefix, these models are trained to predict the activity instance(s) that is/are most likely to occur next in the case. Given this next-activity prediction model, these techniques predict the case suffix by repeatedly applying the model to predict the next activity instance, the following one, and so on, until the model predicts the end of the case.    

A recognized shortcoming of these approaches is their tendency to generate suffixes that contain an excessive number of consecutive (repeated) instances of the same activity, even when such repetitions do not occur or rarely occur in the historical execution data ~\cite{TaxVRD17,KendallGC18,0001DR19,RamaManeiroVL23}. These self-repetition patterns greatly affect the performance of these techniques, both with respect to their ability to replicate the control-flow of the process (sequence of activities) and the temporal behavior (the completion times)~\cite{KendallGC18}. 

In this setting, this paper focuses on the following problem: 
\begin{quote}
    Given an ongoing (incomplete) case of a process, of which we know the sequence of activity instances up to a point in time (herein called a \emph{case prefix}), predict the sequence of activity instances until completion of the case and the end timestamp of each activity instances (herein called the \emph{case suffix}) while avoiding generating activity repetitions that are not present in the historical process execution data.
\end{quote}


To reduce or avoid activity repetitions not present in the historical process execution data, it has been proposed to use random selection\cite{0001DR19} of the next activity, instead of always selecting the next activity with the highest probability\cite{TaxVRD17}.
This approach, however, does not consider historical activity patterns in a case when predicting the next activity. As a result, these models end up in an ``exploitative cycle'', where they favor familiar patterns (an activity is repeated), instead of exploring a more diverse set of possibilities.


In this paper, we introduce a  sampling approach to select each subsequent activity in the case suffix. The goal of the proposed approach is to generate case suffixes that more accurately reflect those found in the historical execution data. To achieve this, our approach employs an exploration heuristic. This heuristic ensures that every activity in the actual suffix is represented at least once in the predicted suffix, thus balancing exploration (searching for less frequent activities) and exploitation (favoring the most frequent activities). Such a balance enables the predictive model to explore new or less common activities while still taking advantage of what has been successful in the past. The paper reports on experiments designed to evaluate the performance of the proposed approach relative to baseline approaches in the context of real-life event logs.



The rest of the paper is structured as follows. Section \ref{section2} provides background and a review of related work. In Section \ref{section3}, we present our proposed approach. Section \ref{section4} offers describes the datasets, evaluation metrics, baseline methods, and the experimental setup. Our findings are presented in Section \ref{section5}, followed by a discussion on threats to validity in Section \ref{sec:discussion}. Finally, Section \ref{section7} concludes the paper and highlights areas for future research.

\section{Background and Related Work}\label{section2}


Existing approaches for predicting activity sequences of business processes generally relies on neural network architectures. These approaches can be categorized according to the neural network architecture employed: Long Short-Term Memory (LSTM), Recurrent Neural Networks (RNN), Convolutional Neural Networks (CNN), and Generative Adversarial Networks (GAN).

LSTMs\cite{GersSC00} are a type of RNN capable of learning and remembering information across long sequences. They are well-suited for time-series data like process monitoring due to their ability to capture long-term dependencies and their ability to seamlessly handle  sequences of varying lengths\cite{WangYLS19}. 
CNNs \cite{SimonyanZ14a}, primarily known for image recognition tasks, have been adapted for sequence data through the use of 1D convolutions. They can capture local patterns in data which can be useful for certain process monitoring tasks. GANs \cite{TaymouriREBV20} consist of two networks, a generator, and a discriminator, that are trained together. For process monitoring, they can be utilized to generate synthetic process data or to learn and differentiate between regular and anomalous process sequences. With these approaches in mind, the following section discusses related works that have delved into their application in predictive process monitoring.


Tax et al.\cite{TaxVRD17} highlighted the potential of LSTMs by demonstrating their capability in predicting not only the next event of a process, but also the entire case suffix and remaining time. Di Francescomarino et al.\cite{Francescomarino17} extend the approach in \cite{TaxVRD17} by leveraging both the inherent structure of process traces and any a-priori knowledge. Their work showed improved accuracy compared to standard LSTM models. 
Other authors have proposed alternative LSTM architectures that improve the prediction of the remaining time of a case~\cite{VerenichDRMT19,GunnarssonBW23}



Lin et al.~\cite{LinWW19} emphasized the importance of event attributes in predictions. Their RNN based model, equipped with a component modulator, effectively integrated multiple event attributes, exhibiting enhanced prediction capabilities on several real-life datasets.


While the LSTM architecture is the most commonly studied architecture for activity sequence prediction, Mauro et al.~\cite{MauroAB19} explored the potential of convolutional neural networks (CNNs) in the same domain. Using stacked inception CNN modules, their findings suggest that CNNs could offer advantages in efficiency and accuracy for next-activity prediction tasks compared to some RNN-based approaches. Along a similar line, Pasquadibisceglie et al.~\cite{Pasquadibisceglie20} introduced a technique that employs CNNs to predict upcoming activities in a trace. This method transforms historical event log data into 2D spatial constructs, similar to images. When training the CNN on these constructs, a deep learning model is developed, aimed at proactively predicting the next activity. However, while this model seems effective at forecasting the next event, it appears to have challenges in predicting the associated timestamp.


Taymouri et al.~\cite{TaymouriREBV20} discuss shortcomings of LSTM and CNN approaches for activity sequence prediction and address these shortcomings by proposing an adversarial training framework adapted from Generative Adversarial Networks (GANs). This approach pits two neural networks against each other, generating case suffixes that more closely resemble the ones in the ground truth. 


A common thread in the above approaches is that they train neural network models to predict the next activity instance in a case. They then use this predictor in an auto-regressive manner. In other words, given a trace of an ongoing case (a case prefix), they start by predicting the next activity instance. Next, given the case prefix plus the predicted next activity instance, they predict the subsequent activity instance, and so on, until they predict the end of the case.
At each step, these approaches internally predict multiple possible next activities, with different probabilities. Among the possible next activities, the above approaches select the most likely one as the next activity for a case suffix. This approach often results in predicting the most frequent activity from a set of possible next activities for a sequence. This exploitative nature leads to spurious activity repetitions in the predicted case suffix. Camargo et al.~\cite{0001DR19} demonstrate that selecting the next activity randomly using the probabilities produced by the next-activity predictor, rather than always selecting the most probable one, enhances the accuracy and quality of the predicted case suffixes. This random selection approach is fundamentally explorative -- it considers less probable options at each step alongside more probable ones. Building on this, the present paper aims to propose an approach that strikes a balance between exploration and exploitation for selecting the next activity in a case suffix. 

\section{Daemon Action Approach}\label{section3}


In line with the previous discussion, we propose an approach for selecting the next activity in a case suffix from the most likely activities suggested by the next activity predictors. This approach is designed to strike a balance between exploration and exploitation. To achieve this, we introduce a `Daemon Action' approach, which aims to refine the process of sampling the next activity for a case suffix. This method focuses on precisely choosing from the most likely activities, ensuring a more targeted and effective balance between the two strategies.



Inspired by the Ant Colony Optimization Algorithm\cite{Dorigo}, Daemon Action enable refining solutions and directing the search process. In computing, a daemon typically refers to a background process. Within the context of Ant Colony Optimization, daemon actions serve as supplementary processes that guide and fine-tune the search to produce more accurate results.


Equation \ref{da} shows the proposed Daemon Action approach formulated as a mathematical equation and can be integrated  with existing deep learning approaches for case suffix and remaining time prediction.\\

\begin{equation} 
F(a) = \frac{{P(a) \cdot \frac{1}{{\text{{count}}(a)}}}}{{\sum_{i=1}^{N} \left( P(i) \cdot \text{{count}}(i) \right)}}
\label{da}
\end{equation}\\

Here, $F(a)$ represents the function that determines the selection of the next activity in a suffix from a sample. The term $P(a)$ signifies the probability of selecting activity $a$ among possible next activities. The term $\text{{count}}(a)$ represents the count of historical occurrence of activity $a$. The summation term $\sum_{i=1}^{N} \left( P(i) \cdot \text{{count}}(i) \right)$ calculates the overall preference or likelihood for any activity based on their respective probabilities and historical occurrences.

The purpose of the Daemon Action approach is to balance between  exploration and exploitation in selecting next activity in a suffix. Exploration refers to the attempt to discover new or less frequent activities, rather than relying solely on what has been successful in the past. In equation \ref{da} the term ($1/count(a)$) encourages exploration. The $count(a)$ represents the number of times an activity $a$ has been selected in the past. If an activity has been selected less frequently (count(a) is low), the reciprocal $1/count(a)$ will be higher. By multiplying the probability $P(a)$ by this reciprocal, activities historically less frequent are given more weight in the calculation of $F(a)$. This means that the equation will tend to favor less common activities, encouraging the model to explore new or less frequent activities.

Exploitation refers to using strategies that were successful in the past. It leverages known paths or activities that previously showed optimal results, termed as the \textit{most promising areas} in the search space. The algorithm minimizes redundant exploration by building on past insights. In equation \ref{da}, the term $P(a)$ encourages exploitation. This term represents the likelihood of selecting an activity among all available activities in a sample. If an activity has a higher probability, it indicates that the model has found this activity to be more frequent in the past. Hence, activities with higher probabilities (or higher historical success rates) are given more weight in the calculation of $F(a)$.

The equation \ref{da} offers a dynamic trade-off between exploration and exploitation by balancing the two conflicting goals. It achieves this by multiplying the probability $P(a)$ by the reciprocal of the count $1/count(a)$, and then normalizing by the sum over all activities. By combining these values, the Daemon Action approach allows the model to explore new or less common activities while still taking advantage of what has been successful in the past and considering the probabilities to guide decisions about which activity to include in a case suffix. Hence, this balance can lead to more effective and robust predictive modeling.

\section{Experiment Design}\label{section4}

This section reports on an evaluation of the proposed daemon action approach via computational experiments on real-life event logs. The objective of this evaluation is to compare the accuracy of the Daemon action approach, layered on top of different LSTM architectures for next-activity prediction, relative to other sampling approaches. 

\subsection{Questions}
Given that the targeted problem is that of predicting sequences of activity instances, such that each activity instance consists of the activity type and the timestamp, we are interested in assessing the relative accuracy of the proposed approach both with respect to the control-flow perspective (the order in which the activity types appear in the predicted sequence) and with respect to the temporal perspective (the accuracy of the timestamps, particularly the last timestamp in the predicted sequence). Additionally, since the Daemon action approach is designed to prevent the occurrence of spurious activity repetitions in the predicted suffix, we are also interested in assessing if the repetition patterns produced by Daemon action are more reflective of the ground truth than those produced by baseline approaches. 
Accordingly, the experiments presented herein address the following Experimental Questions (EQs):
\begin{itemize}
\item[EQ1] Does the proposed Daemon Action approach, combined with different LSTM architectures for next-activity prediction, predicts sequences of activity types that are closer to the ground truth, compared to other sampling approaches?
\item[EQ2] Does the proposed Daemon Action approach, combined with different LSTM architectures for next-activity prediction, predicts case suffixes with a distribution of activity repetitions closer to the ground truth, compared to other sampling approaches?
\item[EQ3] Does the proposed Daemon Action approach, combined with different LSTM architectures for next-activity prediction, predicts case suffixes with an overall temporal accuracy closer to the ground truth, compared to other sampling approaches?
\end{itemize}



\subsection{Datasets}\label{datasets}
We conducted our experiments using 15 real-life event logs. To ensure the reproducibility of our experiments, the logs employed are publicly accessible at the \textit{4TU Centre for Research Data}\footnote{\url{https://data.4tu.nl/search?search=bpi}} as of November 2023. Notably, several of these logs originate from various years of the Business Process Intelligence Challenges.

In our evaluation, we excluded logs that presented extreme outliers, such as BPIC 2011 log contains timestamps with a granularity of a day. As a result, there is a large number of activity instances that cannot be chronologically ordered because they fall under the same day (tie-break). The ability to chronologically order the activity instances is a pre-requisite for computing case suffixes. Additionally, logs that did not feature both case attributes (static) and event attributes (dynamic) were excluded, leading to the removal of BPIC 2014. BPIC 2016 was also excluded as it consists of a click-stream dataset from a web service rather than a business process event log. Furthermore, the BPIC 2018 event log was removed from our analysis because of its strong seasonal patterns that make predictions trivial. Specifically, 76\% of the cases in this log terminated on one of three specific days: 2018-01-06, 2017-01-07, and 2016-02-19. Similarly, we excluded BPIC 2019 because most of its cases are very short, with 75\% having a trace length up to 6.

Table \ref{table1} summarizes the characteristics of each resulting dataset. For each dataset, $\#Cases$ indicates the total number of cases, or process executions. The $\#Activity Instances$ column specifies the total occurrences or executions of various activities in a process, while $\#Activity$ denotes the count of unique activities. The $Avg.Act/Trace$ column displays the average number of activities per trace. The columns $Avg. Dur.$ and $Max Dur.$ depict the average and maximum cycle time—the duration from the start of a process execution to its end. Further details of each  dataset are as follows:

\begin{itemize}


 \item BPI12\footnote{doi: 10.4121/uuid:3926db30-f712-4394-aebc-75976070e91f}: This dataset comprises sequences from a loan application process at a Dutch financial institution, captured through an online system between October 1, 2011, and March 14, 2012. Within this process, there are three distinct sub-processes, one of which is labeled as W. Consequently, we isolate this sub-process from the dataset, termed BPI12w.

\item BPI13\footnote{doi: 10.4121/uuid:500573e6-accc-4b0c-9576-aa5468b10cee}:These event logs feature real-life data from Volvo IT's VINST system, focusing on incident and problem management processes. we utilized the closed problem set of event log from the BPI Challenge 2013.

\item BPI15\footnote{doi: 10.4121/uuid:31a308ef-c844-48da-948c-305d167a0ec1}: This dataset comprises event logs from five Dutch municipalities related to the building permit application process, with each municipality's dataset treated as a separate event log.

\item BPI17\footnote{doi: 10.4121/uuid:5f3067df-f10b-45da-b98b-86ae4c7a310b}: The dataset comprises sequences from a loan application process at the same Dutch financial institute as BPI12, covering 2016 and extending up to February 2, 2017.

\item BPI20\footnote{doi: 10.4121/uuid:52fb97d4-4588-43c9-9d04-3604d4613b51}:The dataset includes two years of travel expense claims from two university departments, covering documents like declarations and payment requests, all following a similar process. Each document has been treated as a separate event log.

\end{itemize}

\begin{table}
\centering
\caption{Descriptive Statistics of Datasets}
\label{table1}
\resizebox{\textwidth}{!}{%
\begin{tabular}{>{\raggedright\arraybackslash}p{4.8cm}ccccccc}
\toprule
\rowcolor{headercolor}
\textbf{Eventlog} & \textbf{\#Cases} & \textbf{\#Activity Instances} & \textbf{\#Act.} & \textbf{Avg. Act/Trace} & \textbf{Max. Act/Trace} & \textbf{Avg. Dur.} & \textbf{Max Dur.}  \\
\midrule
BPI2012 & 13087 & 164505 & 23 & 12.57 & 96 & 8.61 days &91.32 days  \\
BPI12w & 9658 & 72413 & 6 & 7.5 & 74 & 11.41 days & 91.02 days  \\
BPI2013 & 1487 & 6660 & 7 &4.48  & 35 &  178.99 days & 6 years, 62 days  \\
BPI2015\_1 & 1199 & 27409 &38  & 22.8 & 61 & 95.9 days &4 years, 26 days   \\
BPI2015\_2 & 832 & 25344 & 44 & 30.4 & 78 &160.3 days& 3 years, 230 days \\
BPI2015\_3 & 1409 & 31574 &40  & 22.4 & 69 & 62.2 days & 4 years, 52 days \\
BPI2015\_4 & 1053 & 27679 & 43 & 26.2 & 83 & 116.9 days & 2 years, 196 days   \\
BPI2015\_5 & 1156 & 36234 & 41 & 31.3 & 109 & 98 days & 3 years, 248 days   \\
BPI2017 & 31509 & 561671 & 26 &  17.83& 61 & 21.84 days & 168.94 days   \\
BPI2017(W) & 31500 & 128227 & 8 &4.07 & 20 & 11.27 days & 167.72 days   \\
BPI20(Domestic\_Declaration) & 10500 & 56437  & 17 & 5.37 & 24 &  11.55 days & 1 year, 102 days  \\
BPI20(International\_Declaration) &6449  & 72151  &34  & 11.19 & 27 & 86.45 days & 2 years, 11 days  \\
BPI20(Permit\_Log) & 5417 & 64107 & 51 & 11.83 & 90  & 86.75 days & 3 years, 95 days   \\
BPI20(Prepaid\_Travel\_Cost) &2099  & 18246  & 29  & 8.69 & 21 & 36.83 days & 325.10 days  \\
BPI20(Request\_For\_Payment) &6886  & 36796 &19  &5.34 & 20 & 12.04 days & 1 years, 40 days  \\

\bottomrule
\end{tabular}
}
\end{table}

\subsection{Experiment Setup}\label{setup}


We implemented the proposed approach in Python 3.8, by leveraging a previous implementation of an LSTM approach for next activity and case suffix prediction reported in~\cite{0001DR19}. We extended this existing implementation with an implementation of the proposed Daemon action approaches as well as other baseline sampling approaches discussed below. The resulting Python scripts as well as instructions on how to reproduce the reported experiments can be found at \href{https://github.com/AwaisAli37405/LSTM---Daemon-Action.git}{GitHub Repository} \footnote{\url{https://github.com/AwaisAli37405/LSTM---Daemon-Action.git}}


\paragraph{Data splitting and hyperparameter optimization} 
To simulate real-life scenarios where models are trained on historical data and tested on current cases, we treat each training and test instance as a pair consisting of one prefix and one suffix event, denoted as $(\sigma \leq k , \sigma > k)$, where the prefix length $k$ is at least 1. We utilized a temporal split\cite{VerenichDRMT19} for dividing the event log into training and testing sets. This process orders the cases by their start time, allocating the first 80\% for training and the remaining 20\% for evaluating predictive accuracy.

Hyperparameter optimization is a crucial step for enhancing model performance. We achieve this by dividing the training set further into 80\% for training and 20\% for validation. This allows us to fit the predictors using different hyperparameter combinations and evaluate their performance on the validation set. The Mean Absolute Error (MAE) serves as the metric for assessing each combination on the validation data. The best-performing combination, as determined by the MAE, is then used to retrain the model on the full training set. The hyperparameter oprimization ranges have been shown in Table \ref{hyp-ranges}. Moreover, to comprehensively explore the hyperparameter space for each dataset, we implement a random search with 50 iterations, focusing on different hyperparameter combinations. This approach ensures a thorough and effective training process for the predictor on each dataset.

\begin{table}[hbtp]
\caption{Hyperparameter Configuration Space}
\centering
\scriptsize
\rowcolors{2}{mylightgray}{white}
\begin{tabular}{
  |>{\centering\arraybackslash}m{3cm}
  |>{\centering\arraybackslash}m{4.8cm}
  |>{\centering\arraybackslash}m{4cm}|}
\hline
\rowcolor{mygray}
\textbf{Parameter - LSTM} & \textbf{Explanation} & \textbf{Search Space} \\ \hline
Batch size & No of Samples to be Propagated & [32,64,128] \\ \hline
Normalization method & Preprocessing - Scaling & [lognorm, max] \\ \hline
Epochs & Number of training epochs & 100 \\ \hline
N\_size &Size of the ngram & [5, 10, 15, 20, 25, 30, 35, 40, 45, 50] \\ \hline
L\_size & LSTM layer sizes & [50,100,150,200,250] \\ \hline
Activation & Activation function to use & [selu, tanh, relu, sigmoid, linear, softmax] \\ \hline
Optimizer & Weight Optimizer & [Nadam, Adam, SGD, Adagrad] \\ \hline
\end{tabular}
\label{hyp-ranges}
\end{table}

\paragraph{Baselines} To integrate the proposed Daemon Action approach into a Deep Learning framework, we considered three LSTM-based neural network architectures\cite{0001DR19} for predicting case suffixes and remaining time in business processes. These architectures are: Specialized, Shared Categorical, and Full Shared. 
The Specialized architecture functions as three independent models without sharing information. The Shared Categorical architecture merges inputs related to activities and roles, sharing the first LSTM layer, aiming to minimize noise from mixing different attribute types (categorical or continuous). The Full Shared architecture combines all inputs and fully shares the first LSTM layer. The evaluation explores which LSTM-architecture best fits the nature of each event log.

To further validate the effectiveness of our approach, we compared the results of suffix and remaining time prediction with other established sampling methods. These include Argmax, Random Choice, Top-K, and Nucleus sampling.

Argmax\cite{0001DR19} sampling methodically selects the next sequence item with the highest probability. In contrast, Random Choice\cite{0001DR19} sampling randomly picks the next item based on the model's probability distribution, introducing greater unpredictability. Top-K\cite{LewisDF18} sampling narrows the selection to the K most probable choices, with the model randomly choosing from these top options. Finally, Nucleus sampling\cite{HoltzmanBDFC20} selects from a dynamically sized pool of top predictions, choosing a subset that collectively reaches a specific probability threshold (p). This approach allows for a more flexible selection range compared to the fixed number in Top-K sampling.

By comparing our method against these diverse sampling approaches, we aim to comprehensively assess its relative performance and accuracy in suffix and remaining time prediction.

\paragraph{Measures of Goodness}

The purpose of \emph{EQ1} is to evaluate the closeness of the sequences of activity types generated by different sampling approaches to the ground truth at the control flow level. To evaluate the proposed daemon action approach  from this perspective, we applied the  established Damerau-Levenshtein distance (DL) metric \cite{Damerau64}.

DL metric enhances the Levenshtein distance by adding a swapping operation, making it more appropriate for evaluating the quality of predicted suffixes. As an illustration, when comparing the sequences \( (a1, a2, a3) \) and \( (a1, a3, a2) \), the metric assigns a cost of 1.0 for the swap between \( a2 \) and \( a3 \).

 For two given activity sequences \( s_1 = fa(\sigma_1) \) and \( s_2 = fa(\sigma_2) \), the similarity is calculated as:
\begin{equation} \label{sdl}
\text{SDL}(s_1, s_2) = 1 - \frac{\text{DL}(s_1, s_2)}{\max(\text{len}(s_1), \text{len}(s_2))}
\end{equation}
Here, \( \text{len}(s) \) denotes the length of sequence \( s \), that is, the count of its elements.
The value of \( \text{SDL} \) ranges between \([0, 1]\). It reaches \(1.0\) when sequences are identical and touches \(0.0\) when the sequences have completely different elements.

For \emph{EQ2}, to evaluate the closeness of the activity repetitions generated by different sampling approaches to the ground truth at the control flow level, we introduce the Repetitive Activity Similarity (RAS) metric defined as:

\begin{equation} \label{ras}
    RAS = 1 - \frac{\sum_{a \in A} |count_{G}(a) - count_{P}(a)|}{\sum_{a \in A}  (|count_{G}(a)| + |count_{P}(a)|)}
\end{equation}

This metric quantifies the variance in activity frequency between a \emph{ground truth} set(G) and a \emph{predicted suffix} set(P).The activity occurrences are counted in both sets to determine the penalty. The process is repeated for each activity. The summation of the penalties is reported as absolute count difference. It is then normalized by the sum of the maximum possible penalties for all pairs in both sets, P and G. A \emph{RAS} score approaching 1 signifies a high predictive accuracy, suggesting that the predicted activities closely mirror the actual ground truth activities. A score near 0 denotes maximum variance in the frequency of activities in both the predicted and the ground truth set.

The \emph{EQ3} is concerned with the closeness of the overall temporal accuracy of the predicted case suffix compared to the ground truth and other sampling approaches. A standard method for evaluating the temporal accuracy of case suffix predictors involves examining the duration's of the case suffix, specifically the time interval between the last activity in the case suffix and the last activity in the case prefix. This measurement is typically made using Absolute Error (AE), which calculates the time difference between the final activity in the case suffix and the final activity in the case prefix. By averaging these AE values across test cases, we report the Mean Absolute Error (MAE) for our assessment. The MAE for Remaining Time can be defined as:

\begin{equation}\label{MAE}
\text{MAE} = \frac{1}{N} \sum_{i=1}^{N} | t_{\text{actual},i} - t_{\text{predicted},i} |
\end{equation}


\section{Results}\label{section5}



This section summarizes the experimental results in relation to the experimental questions outlined in the previous section. The complete set of experiments is available as supplementary material accessible \href{https://figshare.com/s/90177dbff5f4786a11cd}{online} \footnote{\url{https://figshare.com/s/90177dbff5f4786a11cd}}.

The experiments were conducted for each of three LSTM architectures outlined in~\cite{0001DR19}: Specialized Architecture, Shared-Categorical Architecture, and Full-Shared Architecture. For brevity, we do not present the results for every combination of sampling approach (Argmax, Random, Top-K, Nucleus, and Daemon Action) and LSTM architecture. Instead, for each pairing of sampling approach and dataset, we report only the best performance (e.g. the best SDL value) achieved by a given sampling approach on a given dataset, across all three architectures. The detailed results in the supplementary material show that the findings are qualitatively the same irrespective of the LSTM architecture.



\begin{figure}[htbp]
\centering
\begin{minipage}[b]{0.65\textwidth}
  \centering
  \includegraphics[width=\linewidth]{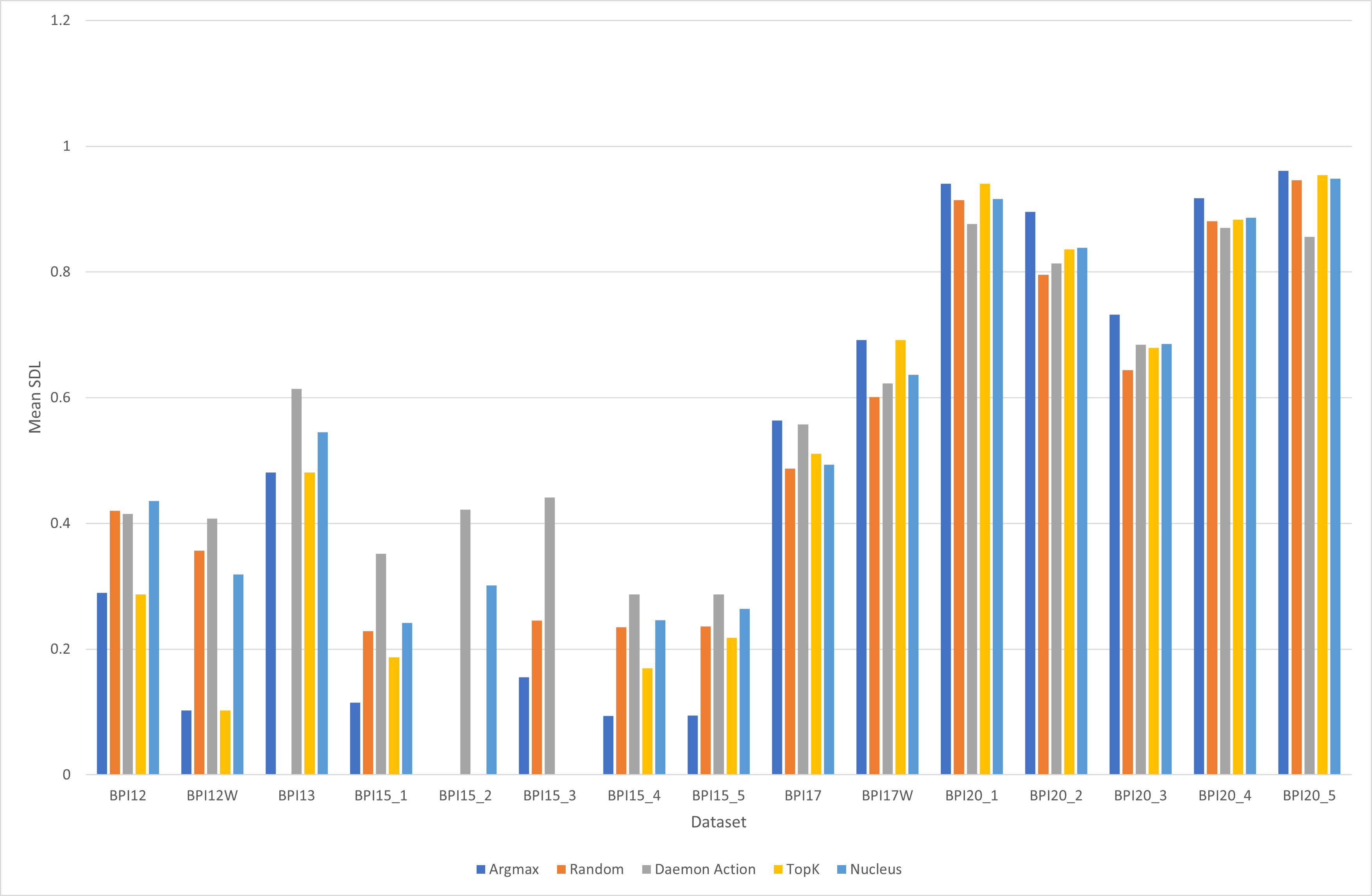}
  \caption{Mean SDL}
  \label{fig:compare_DL}
\end{minipage}\hfill
\begin{minipage}[b]{0.35\textwidth}
  \centering

  \resizebox{\textwidth}{!}{%
  \footnotesize
  
  \begin{tabular}{|l|ccccc|}
  \hline
    \textbf{Dataset} & \multicolumn{5}{c|}{\textbf{Sampling Approaches}} \\
\cline{2-6}
                & \textbf{ArgMax} &\textbf{Random} & \textbf{D-Action} &\textbf{Top-K} & \textbf{Nucleus}   \\
\hline
\hline

BPI12  & 4&	2&	3&	5	&\textbf{1} \\ 
BPI12w& 4&	2&	\textbf{1}	&4&	3 \\ 
BPI13&4	&3	&\textbf{1}	&4&	2 \\ 
BPI15-1&5	&3	&\textbf{1}	&4&	2 \\ 
BPI15-2&5	&4&	\textbf{1}&	2&	3 \\ 
BPI15-3&5	&2	&\textbf{1}	&4	&3 \\ 
BPI15-4&5	&3&	\textbf{1}	&4&	2\\ 
BPI15-5&5&	3	&\textbf{1}&	4	&2 \\ 
BPI17&\textbf{1}	&5	&2&	3&	4  \\ 
BPI17w&\textbf{1}&	5&	4&	\textbf{1}&	3 \\ 
BPI20-1&\textbf{1}&	4&	5&	\textbf{1}&	3 \\ 
BPI20-2&\textbf{1}&	5&	4&	3&	2\\
BPI20-3&\textbf{1}&	5	&3	&4&	2\\
BPI20-4&\textbf{1}	&4	&5	&3&	2\\
BPI20-5&\textbf{1}	&4&	5&	2&	3\\
\hline

  \end{tabular}
}
  
  \captionof{table}{Mean SDL Ranks}
  \label{table-sdl}
\end{minipage}

\end{figure}

\begin{figure}[htbp]
\centering
\begin{minipage}[b]{0.65\textwidth}
  \centering
  \includegraphics[width=\linewidth]{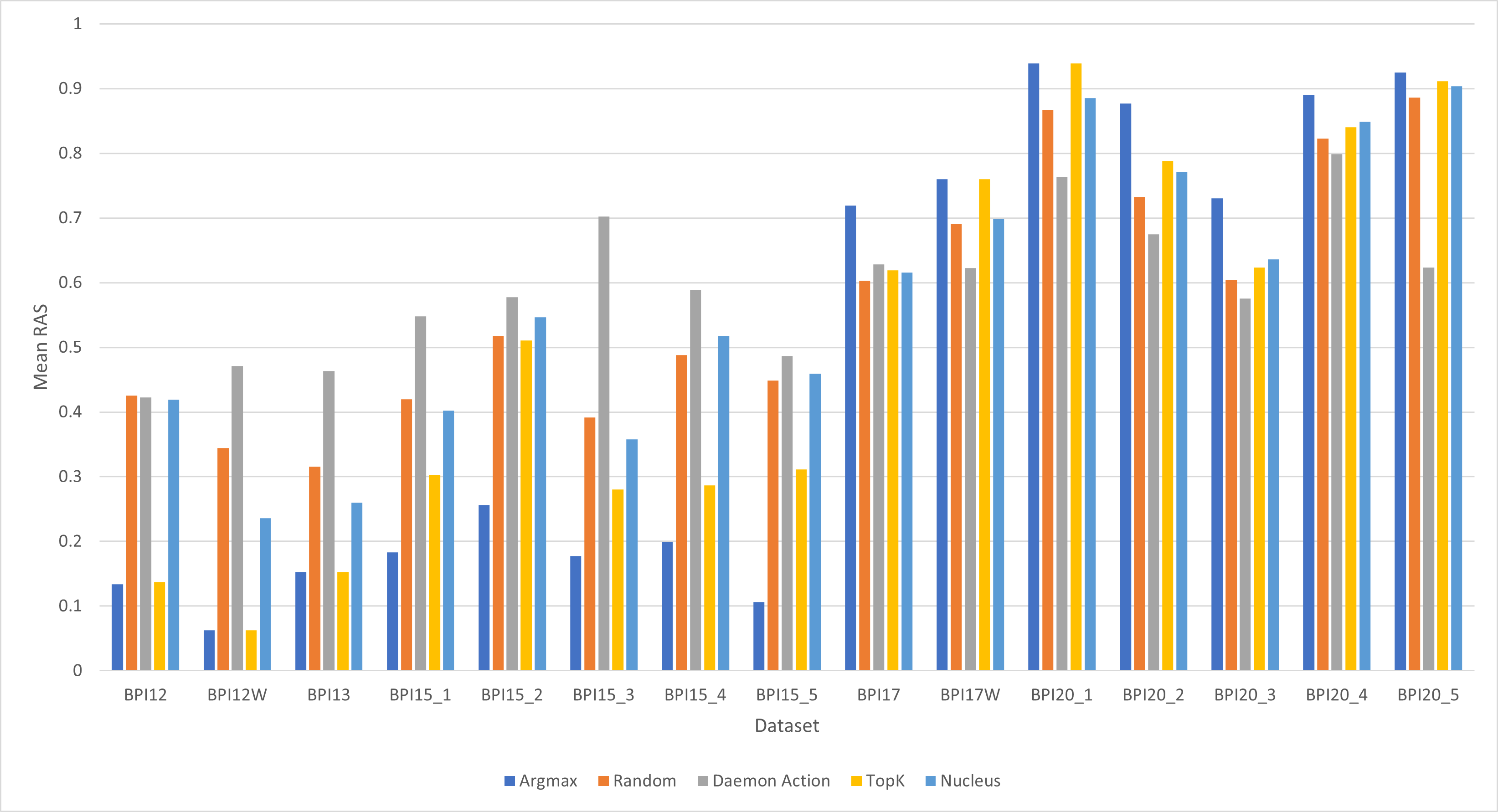}
  \caption{Mean RAS}
  \label{fig:compare_RAS}
\end{minipage}\hfill
\begin{minipage}[b]{0.35\textwidth}
  \centering

  \resizebox{\textwidth}{!}{%
  \footnotesize
  
  \begin{tabular}{|l|ccccc|}
  \hline
    \textbf{Dataset} & \multicolumn{5}{c|}{\textbf{Sampling Approaches}} \\
\cline{2-6}
                & \textbf{ArgMax} &\textbf{Random} & \textbf{D-Action} &\textbf{Top-K} & \textbf{Nucleus}   \\
\hline
\hline

BPI12&5	&\textbf{1}&	2&	4&	3 \\
BPI12w&4&	2&	\textbf{1}	&4	&3\\
BPI13&4	&2&	\textbf{1}&	4	&3\\
BPI15-1&5&	2&	\textbf{1}&	4	&3\\
BPI15-2&5	&3&	\textbf{1}&	4&	2\\
BPI15-3&5	&2	&\textbf{1}&	4&	3\\
BPI15-4&5&	3&	\textbf{1}&	4&	2\\
BPI15-5&5	&3&	\textbf{1}&	4	&2\\
BPI17&\textbf{1}	&5&	2&	3&	4\\
BPI17w&\textbf{1}&	4&	5&	\textbf{1}&	3\\
BPI20-1&\textbf{1}&	4&	5&	\textbf{1}&	3\\
BPI20-2&\textbf{1}&	4&	5&	2&	3\\
BPI20-3&\textbf{1}&	4&	5	&3&	2\\
BPI20-4&\textbf{1}&	4&	5&	3&	2\\
BPI20-5&\textbf{1}&	4	&5&	2&	3\\
\hline
\end{tabular}
}
  \captionof{table}{Mean RAS Ranks}
  \label{table-ras}
\end{minipage}

\end{figure}

\begin{figure}[htbp]
\centering
\begin{minipage}[b]{0.65\textwidth}
  \centering
  \includegraphics[width=\linewidth]{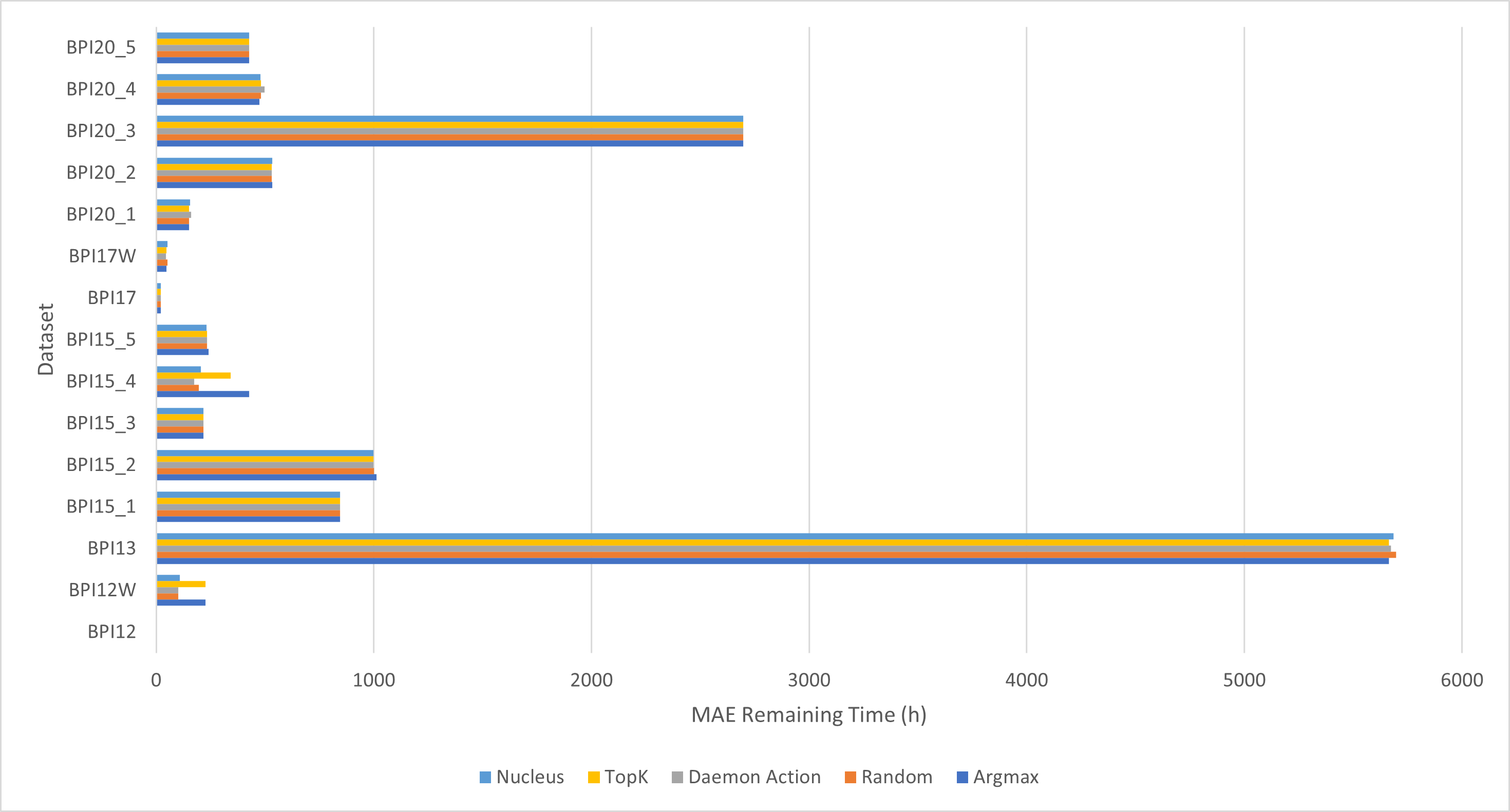}
  \caption{Remaining Time MAE (h)}
  \label{fig:compare_MAE}
\end{minipage}\hfill
\begin{minipage}[b]{0.35\textwidth}
  \centering

  \resizebox{\textwidth}{!}{%
  \footnotesize
  
  \begin{tabular}{|l|ccccc|}
  \hline
    \textbf{Dataset} & \multicolumn{5}{c|}{\textbf{Sampling Approaches}} \\
\cline{2-6}
                & \textbf{ArgMax} &\textbf{Random} & \textbf{D-Action} &\textbf{Top-K} & \textbf{Nucleus}   \\
\hline
\hline

BPI12&\textbf{1}&	\textbf{1}&	\textbf{1}&	\textbf{1}&	\textbf{1} \\
BPI12w&4&	\textbf{1}&	2&	4	&3\\
BPI13&\textbf{1}&	5&	3&	\textbf{1}&	4\\
BPI15-1&5	&4	&2&	\textbf{1}&	3\\
BPI15-2&5&	4&	2&	\textbf{1}	&3\\
BPI15-3&5&	2&	\textbf{1}&	4&	3\\
BPI15-4&5	&2&	\textbf{1}&	4	&3\\
BPI15-5&5	&2&	3&	4&	\textbf{1}\\
BPI17&4	&2&	\textbf{1}&	5&	3\\
BPI17w&2&	5&	\textbf{1}&	2	&4\\
BPI20-1&\textbf{1}	&3&	5&	\textbf{1}&	4\\
BPI20-2&5	&2	&\textbf{1}&	3&	4\\
BPI20-3&\textbf{1}&	\textbf{1}&	\textbf{1}&	\textbf{1}&	\textbf{1}\\
BPI20-4&\textbf{1}&	4&	5	&3&	2\\
BPI20-5&4	&2&	5&	\textbf{1}&	3\\
\hline

  \end{tabular}%
  }
  \captionof{table}{MAE Ranks}
  \label{table-mae}
\end{minipage}

\end{figure}

The results across the three experimental questions are summarized in figures \ref{fig:compare_DL}-\ref{fig:compare_MAE} and tables \ref{table-sdl}-\ref{table-mae}.
Figures \ref{fig:compare_DL}-\ref{fig:compare_MAE} are grouped bar charts plotting the accuracy of each of the five approaches (the Daemon action approach and the four baselines) on each of the datasets. Each group in the chart corresponds to one dataset, and each data series corresponds to one of the approaches. Figure~\ref{fig:compare_DL} plots the SDL metric, Figure~\ref{fig:compare_RAS} the RAS metric, and \ref{fig:compare_MAE} the MAE metric. All the reported values are means across all cases in the test set (e.g.\ Figure~\ref{fig:compare_DL} reports mean SDL values across the cases in the test set).

Next to each grouped bar chart, tables \ref{table-sdl}-\ref{table-mae} provide the ranking of each sampling technique relative to others with respect to each dataset. Table \ref{table-sdl} corresponds to the SDL metric, Table \ref{table-sdl} to RAS and Table \ref{table-mae} to MAE. For example, the cell corresponding to row BPIC12 and column D-Action in \ref{table-sdl} contains a 1, indicating that the Daemon Action (D-Action) sampling approach yielded the highest SDL value on the dataset BPIC12. In some cases, multiple sampling approaches yielded the same value for a given metric, rounded to two decimal points, in which case those approaches get the same ranking (e.g. in Table \ref{table-mae}, all approaches achieved the same MAE for dataset BPI12, so they all have rank 1). 


Regarding \emph{EQ1},  we observe from the Fig. \ref{fig:compare_DL} that SDL Score for the Daemon Action approach generally outperforms other sampling approaches for logs with larger vocabularies (\#Act's) and longer trace lengths, such as  BPI12w, BPI13, BPI15, and BPI17. Likewise, Table \ref{table-sdl} by focusing on Mean SDL scores, demonstrates the strong performance of the Daemon Action approach, which ranks best in 7 out of 15 datasets. This highlights its effectiveness in reducing activity repetitions across a diverse range of datasets. The Argmax approach, though not less consistent, does manage to secure the top position in 7 out of 15 datasets, indicating effectiveness on datasets with shorter trace lengths. Notably, the Nucleus, Top-K and Random Choice approach achieves the best rank on few datasets,  reflecting their limited efficiency in optimizing Mean SDL scores. Hence, this indicates an enhancement in the ability of existing LSTM models by Daemon Action approach to learn traces with longer sequences.

Regarding \emph{EQ2}, it can be seen in Fig.\ref{fig:compare_RAS} and Table \ref{table-ras} that the RAS score of Daemon Action generally surpasses other approaches in reducing repetitions in predicted sequences. Daemon action approach takes into account the historical occurrences of activities in a prefix to predict the next activity in a suffix, as explained in section \ref{section3}. This inherent behavior of Daemon action  makes it  perform better on logs with less repetitive activities in the original sequences such as for logs BPI12, BPI12w, BPI13, BPI15 and BPI17. In contrast, its performance degrades for the logs having high repetitions in the original sequences, for example due to rework loops, such as for logs BPI17w and BPI20. Notably, the Nucleus, Top-K and Random Choice approach again reflects their limited efficiency by performing relatively better on few with respect to Mean RAS score. Whereas, Argmax fails to reduce the redundant activities for datasets with longer trace sequences but performed better on datasets with traces of short length.

As for \emph{EQ3}, Figure~\ref{fig:compare_MAE} shows that the performance of the Daemon action approach is comparable, but arguably not superior, to that of other sampling approaches. This observation is confirmed in Table~\ref{table-mae}, which shows many tie breaks between approaches. While the Daemon Action approach ranks highest in  7 out of 15 datasets in this category, so does the Top-K approach (often ex aequo). The Random Choice, Argmax, and Nucleus methods also have a top rank in several datasets. At first glance, one would expect that higher control-flow accuracy (SDL and RAS metrics) would translate into higher temporal accuracy (MAE of remaining time). However, the results suggest that the predictors are able to correctly predict the remaining time of the case, even if there are errors in the prediction of the sequence of activity labels.


In conclusion, the results collectively reinforce the finding that the Daemon Action approach generally outperforms the Argmax, Nucleus, Random, and Top-K methods in terms of Mean SDL and Mean RAS scores. On the other hand, the results are inconclusive with respect to the MAE scores, which suggests that a separate study is warranted to determine how to take advantage of the improved control-flow predictions, in order to also improve on temporal accuracy.

\section{Threats to Validity}\label{sec:discussion}

We acknowledge certain threats to external and internal validity. External validity refers to the extent by which our results can be extended to other datasets stemming from the use of a limited collection of datasets. To mitigate this threat, we have carefully selected datasets across different domains, sizes, and other characteristics. However, the scope of our datasets may still pose limitations to the broader applicability of our results. Internal validity refers to using one deep learning architecture for sequence prediction, specifically LSTM networks. While this choice is grounded in the demonstrated suitability of LSTMs for similar problems \cite{RamaManeiroVL23}, it might still introduce a potential bias. The exclusive use of LSTMs might limit our understanding of how different architectures could impact the results. Addressing this potential threat to validity is a possible direction for future work.



\section{Conclusion and Future Work}\label{section7}

This paper proposed an approach, namely Daemon action, to sample the next activity instance in an ongoing case of a business process, given a probability distribution of candidate next activity instances produced by a predictive model. The proposed approach aims at balancing between exploitation steps (predicting the most likely activity) and exploration steps (exploring a more diverse set of options) during the generation of case suffixes.

The experimental results show that, when layered on top of different LSTM-based predictors, the Daemon Action approach leads to less occurrences of spurious activity repetitions in the generated case suffixes, which is a known limitation of previous approaches to case suffix prediction.
The experiments also show that the Daemon action approach consistently enhances the control-flow accuracy, measured with respect to the DL distance. In other words, it produces sequences of activity labels that more closely resemble those in the ground truth data.
These improvements are notable when predicting longer sequences of activity instances (i.e. when the approach is applied to business processes with a larger number of activity instances per case).



The experimental evaluation reported in this paper focuses on applying the Daemon Action approach on top of next-activity predictors with LSTM architectures. Future work will aim at evaluating the benefits of the proposed approach when applied to other architectures, such as CNNs and GANs. 


While the Daemon Action approach enhances the  control-flow accuracy (measured via SDL and RAS scores) relative to baseline sampling techniques, we did not observe any notable enhancements in the temporal accuracy (MAE metric). A direction for future work is to design case suffix prediction approaches that explicitly take into account waiting times between activity instances during the generation of the predicted suffixes. This future work would require exploring different types of neural network architectures in combination with different sampling approaches.

\bibliographystyle{splncs04}
\bibliography{ref}




\end{document}